%% file: Main_Text.tex
\documentclass{article}
\usepackage{arxiv}

\usepackage[utf8]{inputenc} 
\usepackage[T1]{fontenc}    
\usepackage{hyperref}       
\usepackage{url}            
\usepackage[dvipsnames]{xcolor}
\usepackage{amsmath}
\usepackage{booktabs}

\usepackage{array}
\usepackage{multirow}

\usepackage{amsfonts}       
\usepackage{nicefrac}       
\usepackage{microtype}      
\usepackage{lipsum}
\usepackage{graphicx}
\usepackage{comment}
\graphicspath{ {./images/} }

\usepackage{float}

\floatstyle{plaintop}
\restylefloat{table}

\usepackage{subcaption}

\def\be{\begin{enumerate}}
\def\ee{\end{enumerate}}

\def\bi{\begin{itemize}}
\def\ei{\end{itemize}}

\def\beq{\begin{equation}}
\def\eeq{\end{equation}}

\def\c2{$\chi^2$}

\newcommand{\etal}{{\it et al.}}

\newcommand{\ie}{{\it i.e.,}}
\newcommand{\eg}{{\it e.g.,}}

\title{Structured Extraction of Process–Structure–Properties Relationships in Materials Science}

\author{
 Amit K Verma \\
  Computational Engineering Division\\
  Lawrence Livermore National Laboratory\\
  Livermore, CA 94550 \\
  \texttt{amitkumar1@llnl.gov} \\
   \And
 Zhisong Zhang \\
  Language Technologies Institute\\
  School of Computer Science\\
  Carnegie Mellon University\\
  Pittsburgh, PA 15213 \\
  \texttt{zhisongz@andrew.cmu.edu} \\
   \And
Junwon Seo\\
  Materials Science and Engineering\\
  Carnegie Mellon University\\
  Pittsburgh, PA 15213 \\
  \texttt{junwons@andrew.cmu.edu} \\ 
  \And
 Robin Kuo \\
  Materials Science and Engineering\\
  Carnegie Mellon University\\
  Pittsburgh, PA 15213 \\
  \texttt{rkuo1@andrew.cmu.edu} \\ 
   \And
 Runbo Jiang \\
  Materials Science and Engineering\\
  Carnegie Mellon University\\
  Pittsburgh, PA 15213 \\
  \texttt{runboj@andrew.cmu.edu} \\ 
   \And
 Emma Strubell \\
  Language Technologies Institute\\
  School of Computer Science\\
  Carnegie Mellon University\\
  Pittsburgh, PA 15213 \\
  \texttt{strubell@cmu.edu} \\
  \And
 Anthony D Rollett \\
  Materials Science and Engineering\\
  Carnegie Mellon University\\
  Pittsburgh, PA 15213 \\
  \texttt{rollett@andrew.cmu.edu} \\
}

\begin{document}
\maketitle
\begin{abstract}

With the advent of large language models (LLMs), the vast unstructured text within millions of academic papers is increasingly accessible for materials discovery—although significant challenges remain. While LLMs offer promising few- and zero-shot learning capabilities, particularly valuable in the materials domain where expert annotations are scarce, general-purpose LLMs often fail to address key materials-specific queries without further adaptation. To bridge this gap, fine-tuning LLMs on human-labeled data is essential for effective structured knowledge extraction \cite{liu_importance_2023}.
In this study, we introduce a novel annotation schema designed to extract generic process–structure–properties relationships from scientific literature. We demonstrate the utility of this approach using a dataset of 128 abstracts, with annotations drawn from two distinct domains: high-temperature materials (Domain I) and uncertainty quantification in simulating materials microstructure (Domain II). Initially, we developed a conditional random field (CRF) model based on MatBERT—a domain-specific BERT variant—and evaluated its performance on Domain I. Subsequently, we compared this model with a fine-tuned LLM (GPT-4o from OpenAI) under identical conditions.
Our results indicate that fine-tuning LLMs can significantly improve entity extraction performance over the BERT-CRF baseline on Domain I. However, when additional examples from Domain II were incorporated, the performance of the BERT-CRF model became comparable to that of the GPT-4o model. These findings underscore the potential of our schema for structured knowledge extraction and highlight the complementary strengths of both modeling approaches.

\end{abstract}


\section{Introduction}

The development of materials through scientific research spans over a century, resulting in a wealth of legacy data embedded in the unstructured text of journals, books, and other sources. 
Using state-of-the-art natural language processing (NLP) methods, including large language models (LLMs), this data—found in text, tables, and figures—can be extracted and transformed into structured databases for alloy development \cite{mysore_automatically_2017, mysore_materials_2019}. 
However, the efficacy of these NLP algorithms depends on domain-specific semantics, which requires manual data tagging, thereby motivating the development of domain-specific ontologies.
A key information retrieval task in this context is named entity recognition (NER), which classifies text tokens into specific categories.
In general-purpose text, these categories typically include names of locations, people, or organizations. 
However, in materials science and engineering, named entities often encompass material-specific terms, such as material properties, characterization techniques, and synthesis \cite{weston_named_2019}. 
Material-specific entities are inherently interconnected—for example, the grain size (a material property) of 100 $\mu$m at a temperature of 1000 $^{\circ}$C under specific environmental conditions. 
Extracting such entities and their relationships as graphs represents another critical information retrieval task \cite{mysore_automatically_2017, mysore_materials_2019}. 
This task involves identifying material-specific entities, extracting relevant relationships, and linking them into graph structures to represent knowledge comprehensively.


Early applications of domain-specific NER in scientific literature primarily focused on extracting drugs and biochemical information to facilitate more effective document searches \cite{rocktaschel_chemspot_2012, garcia-remesal_using_2012}. 
More recently, NER techniques have been adapted to materials science subfields, including inorganic materials \cite{he_similarity_2020}, polymers \cite{tchoua_creating_2019}, and nanomaterials \cite{hiszpanski_nanomaterial_2020}. 
This evolution is comprehensively reviewed in recent literature \cite{olivetti_data-driven_2020, kononova_opportunities_2021}. 
Concurrently, the methodologies employed for NER have progressed from traditional rule-based and dictionary look-up approaches to advanced machine learning (ML) and NLP techniques. 
These include conditional random fields (CRFs) \cite{lafferty_conditional_2001}, long short-term memory (LSTM) networks \cite{hochreiter_long_1997}, and, more recently, transformer-based pre-trained large language models (LLMs) such as BERT \cite{devlin_bert_2018} and GPT from OpenAI \cite{dunn_structured_2022, maikjablonka_14_2023}. 
The accuracy of these models, as measured by precision and recall, has improved significantly, ranging between 60~\% and 98~\%, depending on the complexity of the schema and the size of the annotated dataset \cite{kononova_opportunities_2021}.


Scientific literature often employs domain-specific narratives and language, making it challenging for generic NLP models to extract meaningful information. 
As a result, domain-specific variants of pre-trained language models—such as SciBERT \cite{beltagy_scibert_2019}, BioBERT \cite{lee_biobert_2020}, and MatBERT \cite{walker_impact_2021}—have been developed to capture context-specific concepts and entities. 
However, most existing schemata for NER in materials science are tailored for specific purposes or subdomains, limiting their broader applicability. 
For instance, many studies focus on extracting synthesis recipes due to the absence of fundamental theories predicting outcomes, such as Kim~\etal -exploration of hydrothermal and calcination reactions for metal oxides \cite{kim_materials_2017} or Kononova~\etal- work on solid-state synthesis \cite{kononova_text-mined_2019}.
Although general-purpose LLMs, such as GPT from OpenAI, promise to bridge this gap through few-shot and zero-shot learning, their reliance on vast amounts of general-purpose, unsupervised data presents significant challenges in specialized domains like materials science. 
Human-labeled data remains critical for equipping LLMs with the ability to comprehend the nuanced and complex language of materials science \cite{liu_importance_2023}. 
These annotations not only improve domain alignment but also play an essential role in ensuring the safety, reliability, and accountability of LLM-generated outputs. 

In this study, we developed a general-purpose schema aimed at capturing process-structure-properties relationships for high-temperature structural materials, moving beyond problem-specific schemata published in prior works. 
Additionally, we demonstrate the schema’s applicability to uncertainty quantification within materials science, highlighting its versatility across domains. 
Furthermore, we first train the materials science-specific BERT model (MatBERT) to align our schema with previously published results and then compare its performance to that of a fine-tuned GPT-4o model from OpenAI, evaluating whether general-purpose LLMs can surpass domain-specific BERT models in specialized tasks.

\section{Text Annotation}

To capture the design insights, we focused primarily on paper abstracts, which are typically accessible without any permissions. 
In our experience, we find that most publications report what problem they are addressing (in Red), how they are approaching the problem (in Brown), and what they found (in Green) (sample abstract shown in \autoref{abstract}), which collectively over many abstracts may provide useful design insights.
For the annotation process, relevant publications are manually selected by annotators. 

\begin{figure}[ht]
    \centering
    \framebox{\parbox{\dimexpr\linewidth-20\fboxsep-20\fboxrule}{\small Nickel-based superalloys such as Hastelloy X (HX) are widely used in gas turbine engine applications and the aerospace industry. \textcolor{red}{HX is susceptible to hot cracking, however, when processed using additive manufacturing technologies such as laser powder bed fusion (LPBF).} \textcolor{brown}{This paper studies the effects of minor alloying elements on microcrack formation and the influences of hot cracking on the mechanical performance of LPBF-fabricated HX components, with an emphasis on the failure mechanism of the lattice structures.} \textcolor{OliveGreen}{The experimental results demonstrate that a reduction in the amount of minor alloying elements used in the alloy results in the elimination of hot cracking in the LPBF-fabricated HX; however, this modification degrades the tensile strength by around 140 MPa. The microcracks were found to have formed uniformly at the high-angle grain boundaries, indicating that the cracks were intergranular, which is associated with Mo-rich carbide segregation. The study also shows that the plastic-collapse strength tends to increase with increasing strut sizes (i.e. relative density) in both the ‘with cracking’ and ‘cracking-free’ HX lattice structures, but the cracking-free HX exhibit a higher strength value. Under compression, the cracking-free HX lattice structures’ failure mechanism is controlled by plastic yielding, while the failure of the with-cracking HX is dominated by plastic buckling due to the microcracks formed within the LPBF process.} The novelty of this work is its systematic examination of hot cracking on the compressive performance of LPBF-fabricated lattice structures. The findings will have significant implications for the design of new cracking-free superalloys, particularly for high-temperature applications.}}
    \caption{\normalsize Sample abstract, highlighting the problem in \textcolor{red}{red}, purpose in \textcolor{brown}{brown}, and results in \textcolor{OliveGreen}{green}. DOI of the sample abstract: 10.1016/j.optlastec.2019.105984}
    \label{abstract}
\end{figure}

\begin{figure}[ht]
    \centering
    \includegraphics[width=0.95\textwidth]{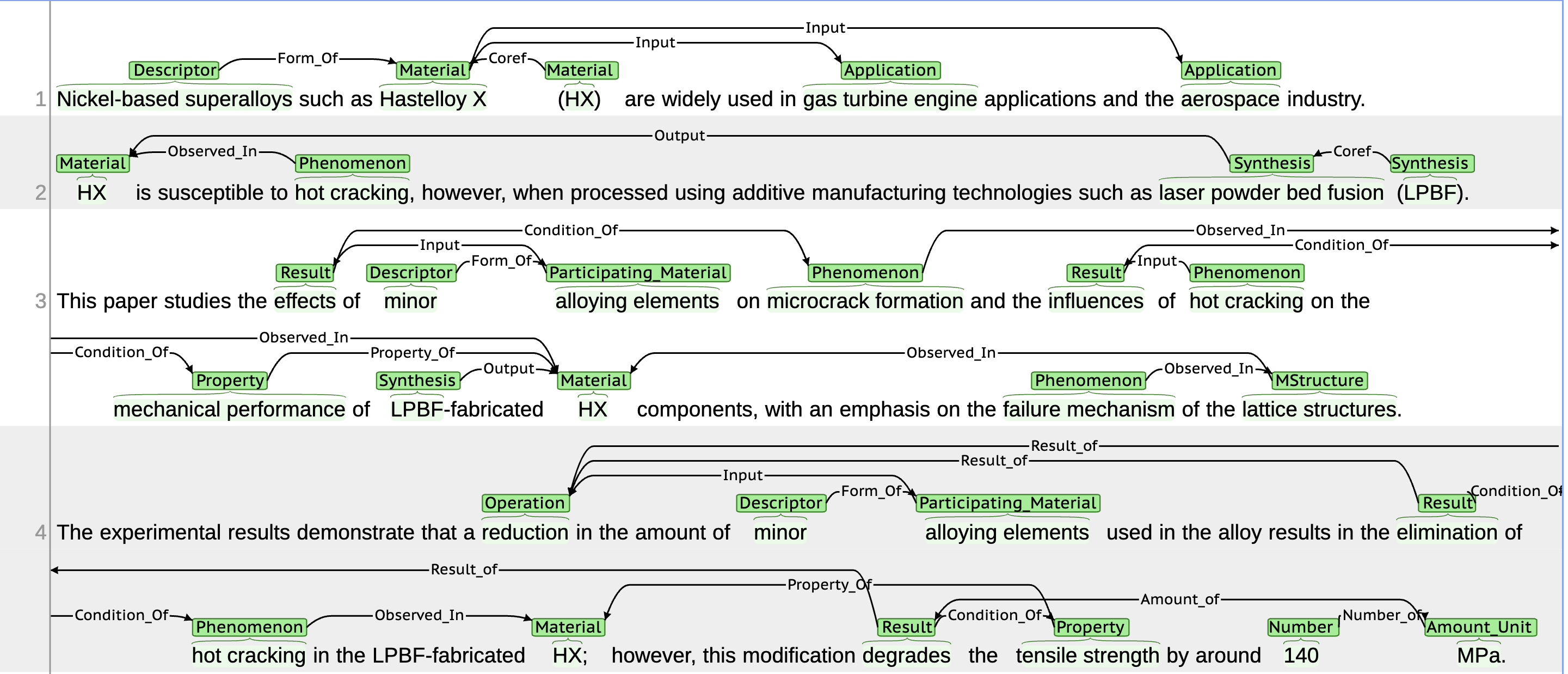}
    \caption{Sample abstract (only first four sentences), after data enrichment in BRAT annotation tool (\url{http://brat.nlplab.org}). DOI of the sample abstract: 10.1016/j.optlastec.2019.105984}
    \label{data_enrich}
\end{figure}

Next, a schema was developed to enrich the data with domain knowledge, and in the process, a training dataset for model development. 
The schema focuses on two aspects: 1) materials science specific entities, and 2) their inter-dependencies. 
Given the focus is on mapping design insights, the entities introduced follow the process - structure - properties loop. 
For example, a \textit{material} sits at the top, its \textit{synthesis}, \textit{microstructure}, \textit{phases}, \textit{properties}, and end \textit{application} defines it, while a specific publication could explore its interaction with an \textit{environment} or a \textit{participating material} to understand a specific underlying \textit{phenomenon} via a single or series of multiple \textit{operation}(s) or \textit{characterization} technique(s). 
The \textit{italicized} concepts in the previous sentence make up the bulk of entities defined in this study, together with a few supporting entities (elaborated in \autoref{entities} with examples).
The key attributes that define these entities are: uniqueness (\ie~no overlap between entities), clarity (\ie~simple enough to be understood by freshman materials science undergraduates), and complementarity (\ie~collectively they can cover a broad range of publications).   
Similarly, inter-dependencies between entities are defined using domain knowledge, elaborated in \autoref{relations} with examples. 
After establishing the schema, the BRAT annotation tool \cite{BRAT} was employed to enrich the data. 
A sample example, after data enrichment, is shown in \autoref{data_enrich}. 

\begin{table}[ht]
\begin{center}
\begin{tabular}{ m{2.5cm} m{6.5cm} m{5cm} } 
  \toprule
  Entity type & Definition & Examples  \\ 
  \toprule
  Material & main material system discussed / developed / manipulated OR material used for comparison & Rene N5 (specific), Nickel-based Superalloy (vague)   \\ 
\midrule
  Participating-Material & anything interacting with the main material by addition, removal, or as a catalyst & Zirconium (mainly elements) \\ 
\midrule
  Synthesis & process/tools used to synthesize the material & Laser Powder Bed Fusion (specific), alloy development (vague)  \\
\midrule
  Characterization & tools used to observe and quantify material attributes (e.g., microstructure features, chemical composition, mechanical properties, etc.) & EBSD, creep test (mostly specific) \\
\midrule
  Environment & describes the synthesis / characterization / operation – conditions / parameters used & temperature (specific), applied stress, welding conditions (vague) \\
\midrule
  Phenomenon & something that is changing (either on its own or as an direct/indirect result of an operation) or observable & grain boundary sliding (specific), (stray grains) formation, (GB) deformation (vague)  \\
\midrule
  MStructure & location specific features of a material system on the ``meso'' / ``macro'' scale & drainage pathways (specific), intersection (between the nodes and ligaments) (vague) \\
\midrule
  Microstructure & location specific features of a material system on the ``micro'' scale & stray grains (specific), GB, slip systems \\
\midrule
  Phase & materials phase (atomic scale) & $\gamma$ precipitate (mostly specific)  \\
\midrule
  Property & any material attribute & crystallographic orientation, GB character, environmental resistance (mostly specific)  \\
\midrule
  Descriptor & indicates some description of an entity & high-angle boundaries, (EBSD) maps, (nitrogen) ions  \\
\midrule
  Operation & any (non/tangible) process / action that brings change in an entity & adding / increasing (Co), substituted, investigate \\
\midrule
  Result & outcome of an operation, synthesis, or some other entity & greater retention, repair (defects), improve (part quality)  \\
\midrule
  Application & final-use state of a material after synthesis / operation(s) & thermal barrier coating \\
\midrule
  Number & any numerical value within the text & 100 \\
\midrule
  Amount-Unit & unit of the number & MPa \\
  \bottomrule
\end{tabular}
\caption{Description of entity types developed in this study, with examples.}
\label{entities}
\end{center}
\end{table}

\begin{table}[ht]
\begin{center}
\begin{tabular}{m{2cm} m{7cm} m{6cm} } 
  \toprule
  Relation & Definition & Examples\\ 
  \toprule
  \multirow{2}{2.5cm}{FormOf} & when one entity is a specific form of another entity; &  single crystal - FormOf - Rene N5  \\ 
  \cline{3-3} 
  & "Descriptor" of "Material" / "Synthesis" / etc. & tertiary - FormOf - $\gamma^\prime$ precipitate \\
  \hline
  \multirow{2}{2.5cm}{ConditionOf} & when one entity is contingent on another entity; & high temperature - ConditionOf - Creep \\
  \cline{3-3} 
  & "Environment" for "Characterization", "Property" for "Phenomenon" & applied stress - ConditionOf - creep test  \\
  \hline 
  \multirow{2}{2.5cm}{ObservedIn} & when one entity is observed in another entity; & GB deformations - ObservedIn - Creep  \\
  \cline{3-3}
  & "Phenomenon" in an "Environment" /or in "Microstructure" /or during "Synthesis" & serrated flow - ObservedIn - tensile deformation \\
  \hline
  \multirow{2}{2.5cm}{PropertyOf} & Specifies where a particular property is found & stacking fault energy - PropertyOf - Alloy3  \\
  \cline{3-3}
  & & environmental resistance - PropertyOf - bond coat \\
  \hline
  Input / Output & Input to an "Operation" / Output of an "Operation"; can be any entity, or a previous operation, or a result of an operation & oxide nanopowders (Input) - 3D extrusion - (Output) extruded filaments  \\ 
  \hline
  ResultOf & connects "Result" with its associated entity / action / operation & suppress (crack formation) - ResultOf - addition (of Ti \& Ni) \\
  \hline
  Next Opr & connects two operations, where one follows the other in the overall process & 3D extrusion - Next Opr - sintering  \\
  \hline
  Coref & link between two description of the same entity, often between the full name and its abbreviation & thermal barrier coating - Coref - TBC \\
  \hline
  Number Of & Designates what unit ("Amount Unit") a "Number" is referring & 700 - Number Of - MPa  \\
  \hline
  Amount Of & Designates what entity a unit ("Amount Unit") is referencing & MPa - Amount Of - applied stress \\
  \bottomrule
\end{tabular}
\caption{Description of relationships between entities.}
\label{relations}
\end{center}
\end{table}

\input{Model_Results}

\section{Fine tuning a large language model}

To compare our BERT-CRF model with commercial off-the-shelf large language models (LLMs), we fine-tuned GPT-4o-2024-08-06 model from OpenAI \cite{noauthor_openai_nodate} using abstracts from Domain I. 
Domain II was excluded from this comparison since its dataset was generated through an active learning approach with the BERT-CRF model. 
This evaluation focused exclusively on the task of extracting entities.
We employed a two-step schema for the named entity recognition (NER) task. 
In the first step, a fine-tuned LLM was used to identify keywords in each sentence. 
In the second step, another fine-tuned LLM classified these extracted keywords. 
The final output was formatted as a JSONL object to encapsulate the details of each entity \cite{dagdelen_structured_2024}. 
The prompts used for fine-tuning were determined by experimenting with zero-shot and few-shot learning approaches; the prompts used in the study are provided in the Supplementary Materials.
For Domain I, the dataset was randomly partitioned into 33 abstracts for training, and 17 abstracts each for development and testing. 
The GPT-4o LLM was subsequently fine-tuned on the training set for both the keyword extraction (step 1) and keyword classification (step 2). 
To evaluate the performance of the fine-tuned model, predictions were repeated three times for each step, resulting in nine predictions per experimental run. 
The entire experiment was repeated three times with different random seeds for the training, development, and test splits, yielding precision, recall, and F1 scores averaged over 27 predictions. These results are presented in \autoref{tab:all} and were subsequently refined in \autoref{tab:error_analysis}.
Our findings indicate that fine-tuning the LLM significantly improved NER performance compared to the BERT-CRF model.
However, when additional examples from both Domain I and Domain II were included, the BERT-CRF model was able to match the performance of the GPT-4o model (\autoref{tab:error_analysis}).
These results highlight the strengths of the BERT-CRF model, which—despite being pre-trained on a smaller dataset—achieved comparable performance to generative LLMs in the context of NER.
This performance advantage is likely attributable to BERT's bi-directional architecture, which provides a more comprehensive understanding of word context compared to the auto-regressive nature of generative LLMs \cite{wang_gpt-ner_2023}.

\section{Error Analysis}\label{sec:error_analysis}

The F1 scores presented in \autoref{tab:all} provide a quantitative assessment of the baseline performance of the models trained in this study. While these scores may appear low for an information extraction task, it is important to note that the exact word-match criterion used for evaluation serves as a conservative lower bound on model performance. In this section, we analyze the types of errors contributing to these scores and examine whether the models successfully extract relevant information even in cases where an exact word match is not achieved. Furthermore, we explore alternative evaluation metrics that, with slight modifications in notation, better reflect model performance in real-world information extraction scenarios.

\begin{table}[ht]
	\centering
	\small
	\begin{tabular}{l | l | l | c c | c c }
		\toprule
		\multirow{2}{*}{Model} & \multirow{2}{*}{Domain} & \multirow{2}{*}{Sentences} &  \multicolumn{2}{c|}{Entity} & \multicolumn{2}{c}{Relation} \\
		& & & dev & test & dev & test \\
		\midrule
        RoBERTa-CRF & I & 533 & $48.96_{0.53}$ & $48.83_{0.23}$ & $52.66_{0.11}$ & $52.85_{1.04}$ \\
	    MatBERT-CRF & I & 533 & $\mathbf{51.67}_{0.61}$ & $\mathbf{52.46}_{0.48}$ & $\mathbf{52.78}_{0.85}$ & $\mathbf{53.73}_{0.62}$ \\
        GPT-4o (both steps) & I & 533 & \textbf{55.51}$_{1.29}$ & \textbf{52.23}$_{1.87}$ & --- & --- \\
        MatBERT-CRF & II & 474 & 59.24$_{0.70}$ & 57.02$_{0.09}$ & 55.09$_{0.75}$ & 55.05$_{1.72}$ \\
        MatBERT-CRF & I \& II & 1007 & \textbf{60.81}$_{1.30}$ & \textbf{57.48}$_{0.45}$ & \textbf{56.96}$_{0.44}$ & \textbf{57.68}$_{1.98}$ \\
        
		\bottomrule
	\end{tabular}
	\caption{Comparison of models developed in this study using F1\% scores for exact match.}
	\label{tab:all}
\end{table}

Here, we categorize entity predictions into five distinct types to capture the nuances of model performance: \textbf{Correct (COR)} entities exhibit exact agreement with the ground truth in both boundary and type, representing \textbf{true positives}. \textbf{Incorrect (INC)} entities have correctly identified boundaries but are assigned an incorrect type, constituting \textbf{classification errors} and thus \textbf{false positives}. \textbf{Partial (PAR)} entities overlap with but do not exactly match the true boundaries, reflecting \textbf{boundary errors} and counted as \textbf{false negatives}. \textbf{Missing (MIS)} entities are present in the ground truth but remain undetected by the model, also contributing to \textbf{false negatives}. Lastly, \textbf{Spurious (SPU)} entities are model-predicted entities that have no corresponding annotation in the ground truth, making them \textbf{false positives}. This taxonomy provides a more granular evaluation of model errors, enabling targeted improvements in both entity recognition and classification.
For the BERT-CRF model, error analysis indicates that partial entity overlap (\textbf{PAR}) is the most prevalent issue, followed by missed entities (\textbf{MIS}), incorrect entities (\textbf{INC}), and spurious entities (\textbf{SPU}) in evaluations for Domain I. However, when considering the entire dataset (Domain I \& II), incorrect entities (\textbf{INC}) surpass missed entities (\textbf{MIS}).
A similar error analysis of GPT-4o results reveals a comparable trend, with partial entity overlap (\textbf{PAR}) being the most frequent issue, followed by spurious entities (\textbf{SPU}), incorrect entities (\textbf{INC}), and missed entities (\textbf{MIS}) in evaluations for Domain I.

The predominance of partial entity overlap (\textbf{PAR}) underscores challenges in precisely delineating entity spans. These challenges may arise due to inconsistent annotation practices, ambiguous boundary cues, or intrinsic limitations in the model’s span detection capabilities.
A closer examination reveals that, in many cases, a single ground truth entity is either split into multiple predicted spans or multiple ground truth spans are merged into a single prediction, while the correct entity type is still identified. Additionally, minor discrepancies in the span boundaries often involve symbols such as $,/(.)$, where one annotation includes the symbol while the other omits it, despite correctly recognizing the entity.
To systematically account for such cases as correct (\textbf{COR}) or true positives, without requiring manual inspection of the entire dataset, we implement an automated string matching approach. Specifically, if one string is fully contained within another, we then verify whether their entity types match. If they do, the instance is updated to a true positive.
Examples illustrating the impact of this adjustment are provided in \autoref{tab:boundary_error}.

\begin{table}[ht]
	\centering
	\small
	\begin{tabular}{p{1.2cm} | p{3.5cm} | p{2cm} | p{4cm} | p{2cm} | p{1.2cm} }
		\toprule
		Model & Entity GT & Type GT &  Entity Predicted & Type Predicted & Corrected \\
		\midrule
	    MatBERT & \textit{investigated} & Operation & \textit{investigated.} & Operation & Yes \\
        -CRF & \textit{HEA}	& Material & \textit{(HEAs)} & Material & Yes \\
         & \textit{Inconel 600} & Material & \textit{Inconel 600 alloy} & Material & Yes \\
         & \textit{electron beam} & Environment & \textit{electron beam melting fusion} & Synthesis & No \\
         & \textit{melting fusion processes} & Synthesis & \textit{electron beam melting fusion} & Synthesis & Yes \\
         & \textit{high temperature} & Environment & \textit{high temperature solar receivers} & Application & No \\
         & \textit{solar receivers} & Application & \textit{high temperature solar receivers} & Application & Yes \\
         & \textit{hree - dimensional} & Descriptor & \textit{three - dimensional} & Descriptor	& Yes \\
        \midrule
        GPT-4o & \textit{density} &	Property & \textit{high-density} & Property & Yes \\
         & \textit{green part} & Descriptor & \textit{green parts} & Descriptor & Yes \\
         & \textit{synchrotron}	& Descriptor & \textit{synchrotron x-ray imaging} & Characterization & No \\
         & \textit{x-ray imaging} & Characterization & \textit{synchrotron x-ray imaging} & Characterization & Yes \\
		\bottomrule
	\end{tabular}
	\caption{Examples of partial entity overlap or boundary errors, with the \textit{corrected} column indicating cases where relaxed criteria are applied to count a match as a true positive.}
	\label{tab:boundary_error}
\end{table}	

The scores after this correction are presented in \autoref{tab:error_analysis}, under the column \textit{Partial Overlap}.
\textbf{Note:} The data for the BERT-CRF model was re-distributed into train, development, and test sets and re-trained for this analysis. Consequently, the scores differ slightly from those in \autoref{tab:all}, which are reported under the column \textit{Current Seed} in \autoref{tab:error_analysis}.
For Domain I, the BERT-CRF model used the same data distribution as the GPT-4o training, utilizing three different seeds. For Domains I \& II, the BERT-CRF model was trained on re-distributed data with seven different seeds.
For the GPT-4o models, we used the same results as in \autoref{tab:all} but updated the scores in the column \textit{Calculation Correction} using the same methodology applied to the BERT-CRF models. This adjustment accounts for a mismatch caused by the strict requirement for an exact match in both entity type and entity text in \autoref{tab:all}, which led to an increased number of both false positives and false negatives.
The dominant error types after this correction for different models are shown in the column \textit{Order after Correction}.

\begin{table}[ht]
	\centering
	\small
	\begin{tabular}{m{1.3cm} | p{1cm} | p{1cm} | p{1cm} | p{1.4cm} | p{1.1cm} | p{1.1cm} | p{1cm} | p{3.2cm} }
		\toprule
		 &  & \multicolumn{4}{c|}{F1\% scores} &  \multicolumn{2}{c|}{Overall}  &  \\
        \cmidrule(lr){3-6}
         Model & Domain  & Baseline & Current Seed &  Calculation Correction & Partial Overlap & Precision & Recall & Order after Correction \\
		\midrule
        MatBERT & I & \textbf{52.5}$_{0.48}$ & \textbf{56.2}$_{2.17}$ & N/A & \textbf{64.9}$_{1.46}$ & \textbf{69.5}$_{2.25}$ & \textbf{61.0}$_{2.75}$ & MIS > INC > PAR > SPU \\
	    -CRF & I \& II & \textbf{57.5}$_{0.45}$ & \textbf{60.2}$_{1.57}$ & N/A & \textbf{72.8}$_{1.20}$ & \textbf{74.0}$_{1.53}$ & \textbf{71.6}$_{1.95}$ & PAR > INC > MIS > SPU  \\
        \midrule
        GPT-4o & I & \textbf{52.2}$_{1.87}$ & N/A & \textbf{62.4}$_{1.42}$ & \textbf{70.5}$_{0.62}$ & \textbf{65.8}$_{0.96}$ & \textbf{76.0}$_{0.76}$ & SPU > INC > PAR > MIS \\
		\bottomrule
	\end{tabular}
	\caption{Scores after applying relaxed text span matching (illustrated in \autoref{tab:boundary_error}) for the entity prediction task.}
	\label{tab:error_analysis}
\end{table}

The results underscore the trade-offs inherent in these approaches. The GPT-4o model appears to excel in recall, capturing more entities overall, yet its tendency toward over-prediction increases the incidence of false positives. In contrast, the BERT-CRF model, when trained on a smaller dataset, seems more conservative—resulting in fewer over-predictions but at the cost of missing some entities. Notably, augmenting the training data for the BERT-CRF framework not only boosts its overall performance but also shifts the error profile toward a more balanced distribution, suggesting that data scale plays a crucial role in mitigating both under- and over-prediction.                                 

After correcting for boundary overlap, classification errors (denoted as \textbf{INC}) emerge as the second most prevalent error type across all models, indicating significant confusion between entity categories. This confusion often arises from overlapping semantic features, such as the similarity between \textit{Microstructure} and \textit{MStructure}.
Although the annotation schema was designed with principles of uniqueness, clarity, and complementarity, these attributes are not consistently maintained across all entities and relations, as discussed below. In particular, context-based annotation introduces variability in entity labeling within the dataset. Manual annotation frequently results in the same token being labeled differently depending on contextual factors. For example, in some abstracts, the phrase ``Ni-based superalloy'' is annotated as a \textit{Material}, whereas in others, ``Inconel 718'' is identified as the primary \textit{Material}, with ``Ni-based superalloy'' classified as its \textit{Descriptor}.
Since some of our models employs contextualized encoders, such as MatBERT, which leverage the surrounding context to generate embeddings, it can distinguish these variations when provided with sufficient training examples. However, such variability adversely impacts test performance, particularly when the context distribution in the test set differs from that in the training set, effectively creating an out-of-domain scenario, as also observed in Section 4.4.
Analysis of the annotated dataset (domain I) reveals that 410 of 3,100 total tokens were annotated inconsistently in different contexts. Notably, approximately half of these tokens were labeled as \textit{Descriptor} (F1-score: 57.75\%) when a more specific annotation was available.
These findings underscore key areas for improvement, including enhanced span boundary detection, more robust contextual embeddings, and refined training data to mitigate annotation ambiguities and improve entity recognition performance.

\section{Discussion}\label{sec:discussion}

The F1 scores reported in \autoref{tab:all} provide an initial overview of the models’ raw performance. As described in \autoref{sec:error_analysis}, partial overlap can significantly reduce scores in entity extraction, even when relevant information is successfully identified. \autoref{tab:error_analysis} demonstrates that employing relaxed matching—using string matching criteria where one entity is contained within another, as illustrated in \autoref{tab:boundary_error}—can lead to improved F1 scores. It is conceivable that manual inspection of a representative subset of predictions could further clarify instances in which the extracted information is correct, despite the comparison methodology designating them as false positives or false negatives. For instance, a case in which ``electron beam'' is annotated as \textit{Environment} in the ground truth, while ``electron beam melting fusion'' is predicted as \textit{Synthesis}, results in a false negative despite both annotation strategies being valid—whether segmenting the text into two entities (“electron beam” and “melting fusion”) or treating them as a single composite entity. Although this discrepancy was not examined in detail, a preliminary calculation that considers partial overlap as correct yields an F1 score exceeding 80\% for both the BERT-CRF (Domains I \& II) and GPT-4o (Domain I) models.

For relation extraction, we trained only the BERT-CRF model using ground truth annotations. Although the aggregated best score of approximately 58\% may appear modest, it is important to note that entities and relations are unevenly distributed within the dataset, as highlighted in \autoref{tab:bd}. Since the dataset was constructed at the abstract level, certain entities and relations occur more frequently than others, reflecting their prevalence in the source material. 
Furthermore, the absence of cross-sentence annotations affects the representation of relation samples. In many cases, cross-sentence relations are approximated by linking entities within the same sentence, which may compromise annotation quality. For example, if a \textit{Phenomenon} is \textit{Observed In} a \textit{Material} that does not appear in the same sentence, the annotator might instead associate the \textit{Phenomenon} with a nearby entity such as \textit{MStructure}, resulting in the relation being recorded as “\textit{Phenomenon} is \textit{Observed In} \textit{MStructure}.” Although this is technically correct, it introduces additional possibilities for the relation classification model to consider, potentially lowering overall performance.
To quantify this effect, we analyzed the annotated data by counting the number of possible combinations for each relation type (\autoref{nocross_sentence}). Our qualitative analysis supports these observations, revealing an uneven distribution of samples across different combinations for several relation types, with the exceptions of \textit{Result Of} and \textit{Property Of}. By categorizing relation types into simple, complex, and infrequent groups (\autoref{nocross_sentence}), we found that simpler relations (e.g., ``number-of'' or ``amount-of'') tend to be more localized and easier to predict, whereas more complex relations (e.g., ``result-of'' or ``condition-of'') often require a comprehensive understanding of the entire sentence or abstract.


\begin{table}[ht]
	\centering
	\small
	\begin{tabular}{c | c c c c | c | c c c c}
		\toprule
		type & \# & \eg/\# & F1(\%) & & type & \# & \eg/\# & F1(\%) & \\
		\midrule
        Number Of & 8 & 7 & 84.38 & Simple & Result Of & 9 & 9.2 & 51.61 & Complex \\
        Coref & 12 & 4 & 83.72 & Simple & Property Of & 10 & 13.2 & 48.05 & Complex \\
        Amount Of & 9 & 4.9 & 75.56 & Simple & Observed In & 47 & 3.8 & 41.80 & Infrequent \\
        Form Of & 22 & 11.4 & 69.39 & - & Input & 39 & 4.6 & 37.50 & Infrequent \\
        Condition Of & 53 & 6.15 & 53.51 & - & Output & 28 & 3.32 & 35.77 & Infrequent \\
		\bottomrule
	\end{tabular}
	\caption{Number of possible combinations (\#) of entities for a given relation type. Here, \eg/\# represent number of samples within the training set per combination, on average (\ie we are not counting number of examples for a given entity pair, individually).  }
	\label{nocross_sentence}
\end{table}

Building on our findings for both entity and relation extraction, several pathways can be pursued to enhance model performance.

\begin{enumerate}
    \item \textbf{Expanding and Balancing the Dataset}: Increasing the dataset size is a straightforward approach that allows the model to learn more robust patterns. As shown in \autoref{tab:new2} and \autoref{fig:al}, dataset expansion can yield noticeable improvements. In addition, addressing the uneven distribution of entities and relations—particularly the imbalance observed in abstract-level constructions—could further refine performance by ensuring that both common and sparse classes are well-represented.
    \item \textbf{Exploring Alternative Model Architectures}: For entity extraction, refining or exploring alternative architectures (e.g., modifications to the BERT-CRF framework) could help mitigate issues such as the performance drop due to partial overlap. In the case of relation extraction, novel architectures that can leverage broader context are needed. Incorporating mechanisms for abstract-level predictions would enable the model to capture cross-sentence relationships, addressing the current limitations where cross-sentence relations are approximated using intra-sentence surrogates.
    \item \textbf{Incorporating Cross-Sentence Annotations}: The absence of cross-sentence annotations currently forces annotators to substitute with local surrogates, which can compromise annotation quality and inflate the number of potential relation combinations. Developing methods to accurately annotate and process cross-sentence relations would reduce this discrepancy and improve model accuracy, especially for complex relation types that require a global context.
    \item \textbf{Reducing Schema Complexity}: Simplifying the annotation schema by reducing the number of entity types and relation combinations can further improve performance. A less complex schema minimizes overlap between entities—thereby decreasing the ambiguity in context-based annotations—and limits the number of potential relation combinations. This, in turn, simplifies the task for the model, leading to more consistent and reliable predictions. As our analysis suggests, simpler relations (e.g., ``number-of'' or ``amount-of``) are inherently easier to predict compared to more complex ones (e.g., ``result-of'' or ``condition-of''), which often require a deeper contextual understanding. One possible direction could be to train separate models for specific entity pairs and their associated relations. This targeted approach could eliminate overlapping annotations and further improve model performance by focusing the learning on narrowly defined sub-tasks.
    \item \textbf{Enhancing Annotation Quality and Consistency}: Manual inspection and refinement of a subset of annotations could help identify systematic errors. For example, cases where partial overlaps cause a mismatch between ground truth and prediction might be better addressed through improved annotation guidelines. This iterative feedback loop would further inform model improvements and contribute to higher F1 scores for both entity and relation extraction tasks.
\end{enumerate}

By combining these approaches—increasing data size, exploring alternative architectures, incorporating cross-sentence predictions, reducing schema complexity, and enhancing annotation quality—we can systematically address the current limitations and significantly improve model performance across both entity and relation extraction tasks.

\section{Conclusion}

In this study, we introduce a novel schema for extracting generic process–structure–properties relationships, employing a BERT-CRF architecture on a corpus of 128 abstracts annotated by materials science domain experts. The proposed schema demonstrates versatility across two distinct domains—high-temperature materials (Domain I) and uncertainty quantification in simulating materials microstructure (Domain II). Our experiments reveal that performance varies by entity and relation type, with average F1 scores of 52.5 and 53.7 for Domain I and 57.0 and 55.0 for Domain II. Notably, fine-tuned LLMs (GPT-4o from OpenAI) achieved an entity-level F1 score of 62.4 for Domain I, surpassing the BERT-CRF baseline.
We identified several challenges impacting model performance, including the handling of partial overlaps in entity extraction, the uneven distribution of entities and relations at the abstract level, and the limitations imposed by sentence-level annotations that fail to capture cross-sentence relationships. Our analysis suggests that expanding and balancing the dataset, exploring alternative architectures capable of leveraging broader contextual information, incorporating cross-sentence annotations, and reducing schema complexity are promising avenues for improvement.
This work provides a robust framework and valuable insights for domain experts engaged in literature-based knowledge extraction. Future research will focus on scaling the dataset, utilizing advanced LLMs, and developing dedicated, domain-specific datasets. In addition, moving from sentence-level to abstract-level annotations will be critical for capturing complex relationships more comprehensively. We encourage other researchers to build upon and adapt this schema to further advance the state of knowledge extraction in their respective fields. 
 The complete code for data preprocessing, training of BERT-CRF models, and the manually annotated dataset is publicly available at https://github.com/zzsfornlp/MatIE/.

\section{Supplementary materials}
\subsection{Prompt used to train step 1 LLM}
[\newline
\{'role': 'system', 'content': 'You will be provided with a string, and your task is to extract keywords from it.'\},\newline
\{'role': 'system', 'content': 'The type of each keyword must be one of Material, Participating Material, Synthesis, Characterization, Environment, Phenomenon, Mesostructure or Macrostructure, Microstructure, Phase, Property, Descriptor, Operation, Result, Application, Number, or Amount Unit.'\},\newline
\{'role': 'system', 'content': "'Material' are main material system discussed / developed / manipulated OR material used for comparison"\},\newline
\{'role': 'system', 'content': "'Participating Material' are anything interacting with the main material by addition, removal, or as a catalyst Material"\},\newline
\{'role': 'system', 'content': "'Synthesis' are process/tools used to synthesize the material"\}\newline
\{'role': 'system', 'content': "'Characterization' are tools used to observe and quantify material attributes (e.g., microstructure features, chemical composition, mechanical properties, etc.)"\}\newline
\{'role': 'system', 'content': "'Environment' describes the synthesis / characterization / operation – conditions / parameters used"\}\newline
\{'role': 'system', 'content': "'Phenomenon' are something that is changing (either on its own or as an direct/indirect result of an operation) or observable"\}\newline
\{'role': 'system', 'content': "'Mesostructure or Macrostructure' are location specific features of a material system on the “meso” / “macro” scale"\}\newline
\{'role': 'system', 'content': "'Microstructure' are location specific features of a material system on the “micro” scale"\}\newline
\{'role': 'system', 'content': "'Phase' are materials phase (atomic scale)"\}\newline
\{'role': 'system', 'content': "'Property' are any material attribute"\}\newline
\{'role': 'system', 'content': "'Descriptor' indicates some description of an entity"\}\newline
\{'role': 'system', 'content': "'Operation' are any (non/tangible) process / action that brings change in an entity"\}\newline
\{'role': 'system', 'content': "'Result' are outcome of an operation, synthesis, or some other entity"\}\newline
\{'role': 'system', 'content': "'Application' are final-use state of a material after synthesis / operation(s)"\}\newline
\{'role': 'system', 'content': "'Number' are any numerical value within the text"\}\newline
\{'role': 'system', 'content': "'Amount Unit' are unit of the number"\}\newline
\{'role': 'user', 'content': 'Nickel-based superalloys such as Hastelloy X (HX) are widely used in gas turbine engine applications and the aerospace industry.'\}\newline
\{'role': 'assistant', 'content': "'Nickel-based superalloys', 'Hastelloy X', 'HX', 'gas turbine engine', 'aerospace'"\}\newline
]\newline
\subsection{Prompt used to train step 2 LLM}
[\newline
\{'role': 'system', 'content': 'You will be provided with two strings.'\}\newline
\{'role': 'system', 'content': 'The first string will be a sentence.'\}\newline
\{'role': 'system', 'content': 'The second string will be list of keywords extracted from the first string.'\}\newline
\{'role': 'system', 'content': 'Your task is to identify the type of each keyword in the second string.'\}\newline
\{'role': 'system', 'content': 'The type of each keyword must be one of Material, Participating Material, Synthesis, Characterization, Environment, Phenomenon, Mesostructure or Macrostructure, Microstructure, Phase, Property, Descriptor, Operation, Result, Application, Number, or Amount Unit.'\}\newline
\{'role': 'system', 'content': "'Material' are main material system discussed / developed / manipulated OR material used for comparison"\}\newline
\{'role': 'system', 'content': "'Participating Material' are anything interacting with the main material by addition, removal, or as a catalyst Material"\}\newline
\{'role': 'system', 'content': "'Synthesis' are process/tools used to synthesize the material"\}\newline
\{'role': 'system', 'content': "'Characterization' are tools used to observe and quantify material attributes (e.g., microstructure features, chemical composition, mechanical properties, etc.)"\}\newline
\{'role': 'system', 'content': "'Environment' describes the synthesis / characterization / operation – conditions / parameters used"\}\newline
\{'role': 'system', 'content': "'Phenomenon' are something that is changing (either on its own or as an direct/indirect result of an operation) or observable"\}\newline
\{'role': 'system', 'content': "'Mesostructure or Macrostructure' are location specific features of a material system on the “meso” / “macro” scale"\}\newline
\{'role': 'system', 'content': "'Microstructure' are location specific features of a material system on the “micro” scale"\}\newline
\{'role': 'system', 'content': "'Phase' are materials phase (atomic scale)"\}\newline
\{'role': 'system', 'content': "'Property' are any material attribute"\}\newline
\{'role': 'system', 'content': "'Descriptor' indicates some description of an entity"\}\newline
\{'role': 'system', 'content': "'Operation' are any (non/tangible) process / action that brings change in an entity"\}\newline
\{'role': 'system', 'content': "'Result' are outcome of an operation, synthesis, or some other entity"\}\newline
\{'role': 'system', 'content': "'Application' are final-use state of a material after synthesis / operation(s)"\}\newline
\{'role': 'system', 'content': "'Number' are any numerical value within the text"\}\newline
\{'role': 'system', 'content': "'Amount Unit' are unit of the number"\}\newline
\{'role': 'system', 'content': 'Your answer should be a JSONL file'\}\newline
\{'role': 'user', 'content': 'Nickel-based superalloys such as Hastelloy X (HX) are widely used in gas turbine engine applications and the aerospace industry.'\}\newline
\{'role': 'user', 'content': "'Nickel-based superalloys', 'Hastelloy X', 'HX', 'gas turbine engine', 'aerospace'"\}\newline
\{'role': 'assistant', 'content': '{"Descriptor": ["Nickel-based superalloys"], "Material": ["Hastelloy X", "HX"], "Application": ["gas turbine engine", "aerospace"]}'\}\newline
]\newline


\input{Main_Text.bbl}
\end{document}

%% file: Model_Results.tex

\section{BERT-CRF Model}

We decompose the extraction problem into two sub-tasks: entity extraction and relation classification. 
Following \cite{zhong-chen-2021-frustratingly}, we adopt two separate models to tackle each of them. 
Both models are based on a pre-trained encoder, while the output modeling is specific to the target tasks. 
For the entity task, we adopt a standard linear-chain CRF (Conditional Random Field) layer \cite{lafferty_conditional_2001}, while for the relation task, we adopt a softmax-based classifier layer to judge the relation for each pair of entities. 
We provide more details for the two models in the following paragraphs.

\paragraph{Entity:} We cast entity extraction as a sequence labeling task and utilize the BIO tagging scheme \cite{ramshaw-marcus-1995-text}. The entity module follows a standard BERT-CRF architecture. Assuming that an input sequence of tokens $\{w_1, w_2, \dots, w_n\}$ is given, we feed it to a pre-trained encoder and obtain the contextualized representations for each token $\{h_1, h_2, \dots, h_n\}$. When a token is split into sub-tokens, we 
adopt the representations of the first sub-token \cite{devlin_bert_2018}. Afterwards, we adopt a linear-chain CRF to model the output tag sequence. Specifically, the probability of a tag sequence $T=\{t_1, t_2, \dots, t_n\}$ is given by:
\[
p(T) = \frac{\exp s(T)}{\sum_{T'} \exp s(T')} ~~~~ s(T) = \sum_{i=1}^{n-1} s_T(t_i,t_{i+1}) + \sum_{i=1}^n s_E(t_i)
\]
Following the CRF formalism \cite{lafferty_conditional_2001}, $s_T$ denotes the transition score for nearby tags, where we adopt a transition matrix which is treated as parameters of the model, while $s_E$ is the emission score for each individual token and we 
stack a linear classifier over the output hidden representations. The model is trained with the loss function of negative log-likelihood, which can be efficiently calculated with the forward-backward algorithm. At testing time, we adopt the standard Viterbi algorithm \cite{viterbi1967error} to obtain the most probable prediction.

\paragraph{Relation:} The relation module is similar to the entity module such that the main component is still a pre-trained model for encoding. Nevertheless, the inputs are different. The entity module adopts raw inputs while the relation module further accepts entity markers in the inputs. We specify two markers: types and anchors, which are similar to those in \cite{zhong-chen-2021-frustratingly} but here we directly adding them to the input embeddings. The first marker indicates the entity labels of all the input entities, and we 
assign to each entity type a specific embedding and add the corresponding type embeddings to the inputs. The second marker indicates the position of the entity that we want to attach relations to, and this type of markers is specific to our scoring scheme. In our preliminary experiments, we found that it could achieve obviously better results if one encoding forward pass was focused on a specific entity's relations, that is, it considered only one entity at one time and assessed the relationships with all other entities to this specific one. Therefore, we elected to specify special embeddings as input anchors to enable the model to be aware of the current considered entity. For the output, we adopted a linear classifier to decide the relation $r$ between two entities $e_1$ and $e_2$:
\[
p(r|e_1, e_2) = \text{softmax} (W \cdot [h_{l1}; h_{r1}; h_{l2}; h_{r2}] + b) 
\]
Here, $\{W, b\}$ are the parameters (weight and bias) in the final linear classifier, $[...;...]$ denotes the concatenation operation and ``$l1, r1, l2, r2$'' indicate the positions of the left and right boundaries of the two entities, respectively. The relation module is trained with the standard cross-entropy loss and greedy decoding for each entity pair is adopted in testing.

\section{Main Results}

\subsection{Settings}

\paragraph{Data.} First, we annotate a dataset consisting of 67 abstracts from domain I, \ie~high temperature materials. Our annotation group included one undergraduate students majoring in materials science, who performed the first-pass annotation jobs, and a senior material-science researcher, who performed a second-pass to finalize the annotations. We pre-process the data with the Stanza toolkit \cite{qi-etal-2020-stanza} for sentence splitting and tokenization. This dataset has 533 sentences, around 11.5K tokens and is annotated with 3.1K entities and 3.0K relations. We randomly split the data, where both development and test set contain 17 abstracts each, while the remaining ones were allocated for training. This splits the dataset roughly into 50:25:25 ratio, with 50\% data (33 abstracts) for training, and 25~\% data for development / test dataset. For evaluation, we report labeled and unlabeled F1 scores for both entities and relations. We assume given entities as inputs for the relation task.

\subsection{Results}

\paragraph{Results.} The entity and relation F1 scores are shown in Table~\ref{tab:res}. We run the experiments with three different random seeds and report averaged results with standard deviations. We compare MatBERT with the RoBERTa model \cite{liu2019roberta} which is pretrained on the general domain corpus. In general, MatBERT gives better results, bringing improvements to the RoBERTa based model. This is because MatBERT is trained on a material science corpus of abstracts that are closer to our target domain.

\begin{table}[ht]
	\centering
	\small
	\begin{tabular}{c | c c | c c }
		\toprule
		\multirow{2}{*}{Model} & \multicolumn{2}{c|}{Entity} & \multicolumn{2}{c}{Relation} \\
		& dev & test & dev & test \\
		\midrule
		RoBERTa & $48.96_{0.53}$ & $48.83_{0.23}$ & $52.66_{0.11}$ & $52.85_{1.04}$ \\
		MatBERT & $\mathbf{51.67}_{0.61}$ & $\mathbf{52.46}_{0.48}$ & $\mathbf{52.78}_{0.85}$ & $\mathbf{53.73}_{0.62}$ \\
		\bottomrule
	\end{tabular}
	\caption{Entity and relation results (labeled F1\%).}
	\label{tab:res}
\end{table}

\paragraph{Error breakdowns.} We further provide error breakdowns on the entity and relation types for a single seed, to investigate which types our models are good at and what their main weaknesses may be. 
Table~\ref{tab:bd} shows the results of the MatBERT based models for the types that appear more than 20 times in the dataset. 
It is unsurprising that the simple types, such as ``Amount Unit'' and ``Number'' for entities and ``NumberOf'' for relations can be accurately predicted. 
However, our models still fall behind in more complex types (\eg for entities, the underlying role for a token is a strong function of context; for relations, the number of possible combinations between which a relationship is possible is high) and especially infrequent types (more in \autoref{nocross_sentence}). 
We further explore these scores in section \ref{sec:error_analysis} and in \ref{sec:discussion} in the context of schema design and from an annotator's perspective. 
In the future, we plan to explore more techniques to deal with these low-resource scenarios. 

\begin{table}[ht]
	\centering
	\small
	\begin{tabular}{c c c | c c c }
		\toprule
		\multicolumn{3}{c|}{Entity} & \multicolumn{3}{c}{Relation} \\
		type & F1\% & Test, Dev, Train &  type & F1\% & Test, Dev, Train  \\
		\midrule
	    Characterization & 86.96 & 22, 28, 33  & Number of & 84.38 & 37, 23, 56  \\
	    Number & 80.00 & 35, 23, 56  & Coref & 83.72 & 21, 28, 49  \\
	    Amount Unit & 71.18 & 25, 16, 44  & Amount Of & 75.56 & 25, 17, 44 \\
	    Synthesis & 66.07 & 51, 46, 67 & Form of & 69.39 & 163, 138, 251   \\
	    Phenomenon & 65.75 & 77, 28, 136 &  Condition Of & 53.51 & 214, 132, 326 \\
	    \midrule
		Operation & 40.33 & 59, 71, 96 &  Result Of & 51.61 & 56, 64, 83  \\
		Result & 39.51 & 113, 74, 160 & Property of & 48.05 & 92, 46, 132  \\
		Application & 36.36 & 6, 11, 9  & Observed In & 41.80 & 105, 48, 179  \\
		Phase & 26.92 & 24, 4, 26  & Input & 37.50 & 129, 105, 180  \\
		Microstructure & 20.00 & 18, 12, 23 & Output & 35.77 & 79, 58, 93 \\
		\bottomrule
	\end{tabular}
	\caption{Error breakdowns (labeled F1\%): results on the best and worst five types, along with the number of samples within test, development, and training dataset from one of the three random seeds / runs. }
	\label{tab:bd}
\end{table}

\subsection{Model Comparisons}

We also perform model comparisons against previous work. Specifically, we compare our BERT-based models with those based on ELMo \cite{peters-etal-2018-deep}, which is based on the BiLSTM architecture rather than Transformer. Similar to the BERT cases, we include ELMo models pre-trained both on general English corpora as well as materials science corpora. The latter is provided\footnote{\url{https://figshare.com/s/ec677e7db3cf2b7db4bf}} by \cite{kim2020inorganic}, which is further fine-tuned with 2.5M materials
science journal articles, starting from the standard pre-trained weights. We refer to this fine-tuned ELMo as MatELMo.
We compare different models with the \texttt{annotated-materials-syntheses} dataset\footnote{\url{https://github.com/olivettigroup/annotated-materials-syntheses}} provided by \cite{mysore-etal-2019-materials}. This is a dataset consisting of 230 synthesis procedures annotated by domain experts. Here we perform the mention extraction experiments with different models and the results are shown in Table~\ref{tab:comM}.

\begin{table}[ht]
	\centering
	\small
	\begin{tabular}{c | c c c }
		\toprule
		Model & P(\%) & R(\%) & F1(\%) \\
		\midrule
		ELMo & $72.12_{0.57}$ & $74.08_{1.29}$ & $73.08_{0.50}$ \\
		MatELMo & $75.92_{0.08}$ & $81.78_{0.65}$ & $78.74_{0.27}$ \\
		\midrule
		RoBERTa & $75.97_{1.47}$ & $85.54_{0.08}$ & $80.47_{0.84}$ \\
		MatBERT & $76.66_{0.58}$ & $86.68_{0.30}$ & $81.36_{0.37}$ \\
		\bottomrule
	\end{tabular}
	\caption{Mention extraction test results on \texttt{annotated-materials-syntheses}.}
	\label{tab:comM}
\end{table}

Similar to our previous findings (Table \ref{tab:res}), the models that are pre-trained on materials science data perform better than the general counterparts. Overall, the MatBERT-based model obtains the best results, suggesting the benefits of pretraining on target specific data and the effectiveness of the transformer model.

\subsection{Schema Comparisons}

Comparing our schema with the ones in the \texttt{annotated-materials-syntheses} dataset from previous work \cite{mysore-etal-2019-materials}, we  find that there are many entity and relation types that could potentially be mapped between these two schema. In view of the overlap, a natural question to ask is: do we really need to annotate new data, or is it simply enough to utilize the previous data for our purpose of information extraction?

To show the effectiveness of our newly annotated data, we further perform a schema comparison experiment by considering a transfer-learning method. Specifically, we first manually gather entity and relation label mappings from the \texttt{annotated-materials-syntheses} schema to ours, which are illustrated in Table~\ref{tab:mapE} and \ref{tab:mapR}, respectively. We 
formed this mapping according the closest matching type descriptions from \cite{mysore-etal-2019-materials}. For the types 
that had no clear mappings, we simply discard them from both schema. Generally, our types are more coarse-grained than \texttt{annotated-materials-syntheses}, mainly because of our focus on annotating abstracts where there are fewer details 
and thus little need of types that are too fine-grained. Overall, the mapping rate is general high: we can keep over 90~\% of the entities and 70~\% of the relations from the original \texttt{annotated-materials-syntheses} dataset.

\begin{table}[ht]
	\centering
	\small
	\begin{tabular}{c | c}
		\toprule
		Target (ours) & Sources (\texttt{annotated-materials-syntheses}) \\
		\midrule
		Material & Material \\
		Number & Number \\
		Operation & Operation \\
		Amount-Unit & Amount-Unit, Condition-Unit, Apparatus-Unit, Property-Unit \\
		Descriptor & Material-Descriptor, Apparatus-Descriptor \\
		Environment & Condition-Misc, Condition-Type \\
		Property & Property-Misc, Property-Type \\
		Synthesis & Meta \\
		Characterization & Characterization-Apparatus \\
		\bottomrule
	\end{tabular}
	\caption{Entity label mappings from the schema of \texttt{annotated-materials-syntheses} to ours.}
	\label{tab:mapE}
\end{table}

\begin{table}[ht]
	\centering
	\small
	\begin{tabular}{c | c}
		\toprule
		Target (ours) & Sources (\texttt{annotated-materials-syntheses}) \\
		\midrule
		Next-Opr & Next-Opr \\
		Number-Of & Number-Of \\
		Condition-Of & Condition-Of \\
		Amount-Of & Amount-Of \\
		Form-Of & Descriptor-Of \\
		Input & Recipe-Precursor \\
		Property-Of & Property-Of \\
		Output & Recipe-Target \\
		Coref & Coref-Of \\
		\bottomrule
	\end{tabular}
	\caption{Relation label mappings from the schema of \texttt{annotated-materials-syntheses} to ours.}
	\label{tab:mapR}
\end{table}

We compare models trained on the mapped \texttt{annotated-materials-syntheses} dataset and our filtered ones. We utilize $MatBERT$ model since its provides overall good performance. The evaluation is performed on our test data, since our main goal is to extract relations for data that we are mostly interested at. The results are shown in Table~\ref{tab:mapres}. Generally, the models trained on our data perform much better, especially with better recall scores. There may be two types of mismatches of the previous data within our scenario. Firstly, the mapping is imperfect and there could be label semantic mismatches, that is, although the label names or the type descriptions are similar, there can still be underlying differences with regard to the scope that one type aims to capture. Moreover, there could be domain mismatch, where the source data may not cover all the patterns of the instances that we aim to extract in the target domain. In this way, we show that for our target scenario, our new schema and annotated data are necessary and helpful.

\begin{table}[ht]
	\centering
	\small
	\begin{tabular}{c | c c c | c c c}
		\toprule
		& \multicolumn{3}{c|}{Entities} & \multicolumn{3}{c}{Relations} \\
		\midrule
		& P & R & F1 & P & R & F1 \\
		\midrule
		Mapped & 33.39$_{1.05}$ & 31.28$_{1.09}$ & 32.28$_{0.54}$ & 91.58$_{0.70}$ & 18.65$_{1.11}$ & 30.98$_{1.55}$ \\
		Ours & 54.58$_{0.61}$ & 58.59$_{0.31}$ & 56.51$_{0.44}$ & 72.19$_{0.15}$ & 48.12$_{0.67}$ & 57.75$_{0.45}$ \\
		\bottomrule
	\end{tabular}
	\caption{Evaluation (F1~\%) of models trained with mapped \texttt{annotated-materials-syntheses} and our newly annotated data.}
	\label{tab:mapres}
\end{table}

\section{Adapting to a New Domain}

To verify that our method can be generalized to new scenarios, we further apply our annotation schema and model to a new domain (II) focusing on uncertainty quantification in simulating materials microstructure. Our annotation group included two graduate students majoring in materials science, who performed the first-pass annotation jobs, and a senior materials-science researcher, who performed a second-pass to finalize the annotations. Moreover, we adopt active learning \cite{settles2009active}, which uses the model to select the most ambiguous sentences to annotate instead of annotating the full abstracts. Our annotation process has two stages. In the first stage, we still perform full annotation and annotate all the sentences in each abstract. This stage allows the annotators to become familiar with our schema and also provides seed data for this new domain. This dataset has 34 abstracts with roughly 9K tokens (364 sentences) and is annotated with 2.26K entities and 2.16K relations. In the second stage, we adopt active learning and only annotate the most ambiguous sentences in each abstract. We set the selection ratio to 40~\%, which sets a balance between reducing annotation efforts and capturing the main contents of an abstract. This dataset has 27 abstracts with 275 sentences, where roughly 110 sentences (40~\%) were annotated with 0.97K entities and 0.95K relations. 

\begin{figure*}[ht]
	\centering
	\begin{subfigure}[b]{0.47\textwidth}
		\includegraphics[width=0.975\textwidth]{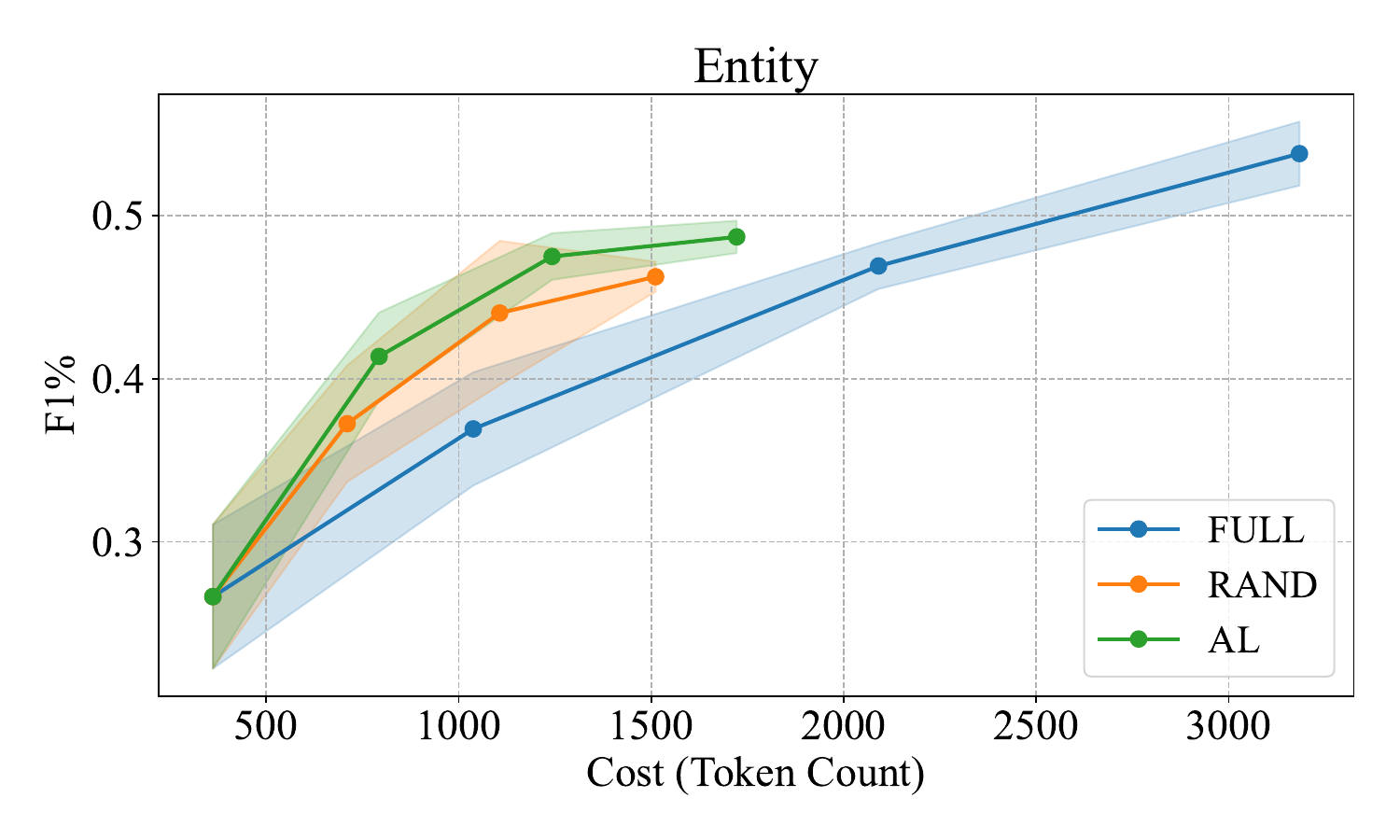}
	\end{subfigure}
	\begin{subfigure}[b]{0.47\textwidth}
		\includegraphics[width=0.975\textwidth]{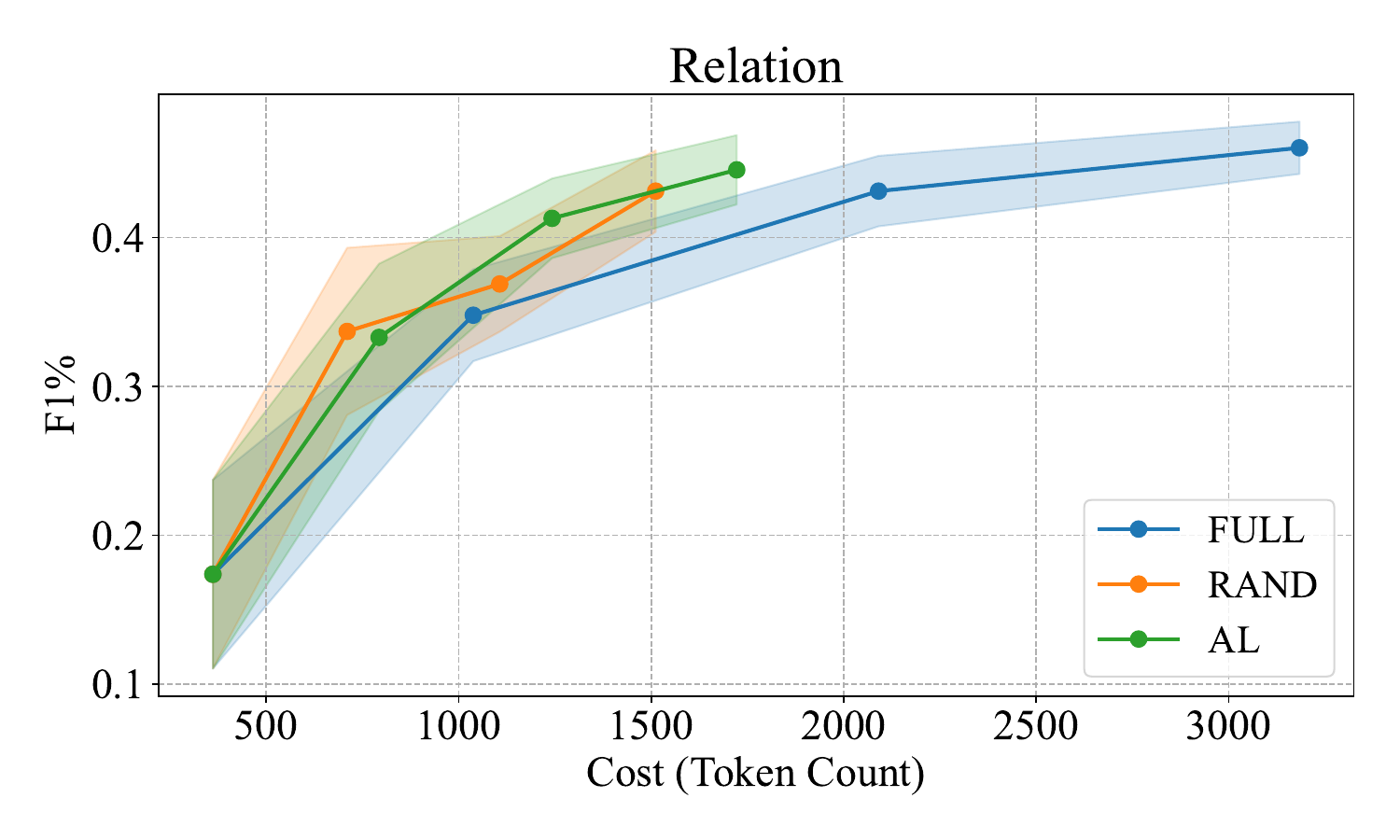}
	\end{subfigure}
	\caption{Comparisons between different sentence selection strategies.}
	\label{fig:al}
\end{figure*}

To investigate the effectiveness of active learning, we take the first subset of our data, which are fully annotated, and evaluate different selection strategies. Specifically, we compare three strategies: full selection (FULL), random selection (RAND), and active selection (AL). In each selection cycle, we pick four abstracts, within which FULL will annotate all the sentences, RAND will randomly choose a subset of sentences (40~\%), while AL will choose the subset by model uncertainty (again 40~\%). The results for both entities and relations are shown in Figure~\ref{fig:al}. Here, the annotation costs (x-axis) is measured by the total token counts in the annotated sentences, since different sentences may have varied lengths and require different annotation efforts. The results show that the AL strategy is the most data-efficient one, and therefore we adopt the AL selection strategy to speed up our annotation process in our second annotation stage.

Table~\ref{tab:new2} shows the main results for this new domain. Here, we also adopt a simple transfer learning scheme by incorporating the annotations from the previous domain into the model training set. The ``Base'' row indicates the results when we only train our models with the fully-annotated abstracts from the first annotation stage, ``+S2'' denotes further addition of the partially-annotated abstracts from the second stage using active learning, and ``+T'' means further using the transfer learning by including annotations from domain I. 
The results suggest that both extra training signals provide benefits for the model performance and with the combination of the two techniques, our models can achieve reasonable performance in the new domain.

\begin{table}[ht]
	\centering
	\small
	\begin{tabular}{l | l | c c | c c }
		\toprule
		\multirow{2}{*}{Model} & \multirow{2}{*}{Sentences} &  \multicolumn{2}{c|}{Entity} & \multicolumn{2}{c}{Relation} \\
		& & dev & test & dev & test \\
		\midrule
		Base & 364 & 55.12$_{0.70}$ & 53.21$_{1.34}$ & 53.83$_{1.13}$ & 51.62$_{0.98}$ \\
        ~~+S2 & 474 & 59.24$_{0.70}$ & 57.02$_{0.09}$ & 55.09$_{0.75}$ & 55.05$_{1.72}$ \\
        ~~+S2+T & 1007 & \textbf{60.81}$_{1.30}$ & \textbf{57.48}$_{0.45}$ & \textbf{56.96}$_{0.44}$ & \textbf{57.68}$_{1.98}$ \\
		\bottomrule
	\end{tabular}
	\caption{Entity and relation results (labeled F1\%) for the new domain.}
	\label{tab:new2}
\end{table}
